\documentclass[runningheads]{llncs}

\usepackage{eccv}

\RequirePackage[width=122mm,left=12mm,paperwidth=146mm,height=193mm,top=12mm,paperheight=217mm]{geometry}

\usepackage[accsupp]{axessibility}  % Improves PDF readability for those with disabilities.

\usepackage[pagebackref,breaklinks,colorlinks,citecolor=eccvblue]{hyperref}
\usepackage{eccvabbrv}
\usepackage{graphicx}
\usepackage{booktabs}
\usepackage{indentfirst}
\usepackage{multicol}
\usepackage{multirow}
\usepackage{makecell}
\usepackage{orcidlink}
\usepackage{marvosym}

\makeatletter
\renewcommand*{\@fnsymbol}[1]{\ensuremath{\ifcase#1\or *\or \dagger\or \ddagger\or
   \mathsection\or \mathparagraph\or \|\or \dagger\dagger
   \or \ddagger\ddagger \else\@ctrerr\fi}}
\makeatother

\begin{document}

% ---------------------------------------------------------------
\title{Learning Unified Reference Representation for Unsupervised Multi-class Anomaly Detection}

\titlerunning{RLR}

\author{
Liren He\inst{1}\thanks{Equal contribution.}\and
Zhengkai Jiang\inst{2*}\and
Jinlong Peng\inst{2*}\and
Liang Liu\inst{2}\and
Qiangang Du\inst{1}\and
Xiaobin Hu\inst{2}\and
Wenbing Zhu\inst{1}\and
Mingmin Chi\inst{1}\textsuperscript{\Letter}\and
Yabiao Wang\inst{3,2}\textsuperscript{\Letter}\and
Chengjie Wang\inst{2}\textsuperscript{\Letter}
}

\authorrunning{L. He et al.}

\institute{
Fudan University, Shanghai, China\\
\email{lrhe21@m.fudan.edu.cn, mmchi@fudan.edu.cn}
\and
Tencent Youtu Lab, Shanghai, China\\
\email{\{caseywang,jasoncjwang\}@tencent.com}
\and Zhejiang University, Hangzhou, China}

\maketitle

\begin{abstract}

In the field of multi-class anomaly detection, reconstruction-based methods derived from single-class anomaly detection face the well-known challenge of ``learning shortcuts'', wherein the model fails to learn the patterns of normal samples as it should, opting instead for shortcuts such as identity mapping or artificial noise elimination. Consequently, the model becomes unable to reconstruct genuine anomalies as normal instances, resulting in a failure of anomaly detection. To counter this issue, we present a novel unified feature reconstruction-based anomaly detection framework termed RLR (\textbf{R}econstruct features from a \textbf{L}earnable \textbf{R}eference representation). Unlike previous methods, RLR utilizes learnable reference representations to compel the model to learn normal feature patterns explicitly, thereby prevents the model from succumbing to the ``learning shortcuts'' issue. Additionally, RLR incorporates locality constraints into the learnable reference to facilitate more effective normal pattern capture and utilizes a masked learnable key attention mechanism to enhance robustness. Evaluation of RLR on the 15-category MVTec-AD dataset and the 12-category VisA dataset shows superior performance compared to state-of-the-art methods under the unified setting. Code is available at \href{https://github.com/hlr7999/RLR}{RLR}.


\end{abstract} 

\section{Introduction}
% 1. background of anomaly detection
Unsupervised anomaly detection and localization strive to learn the patterns of normal samples from the training set then treat outliers as anomalies during inference, which is widely applied in industrial manufacturing~\cite{bergmann2019mvtec, zou2022spot, wang2024real, zhang2024ader}, medical image analysis~\cite{fernando2021deep}, among other fields. However, in practical industrial anomaly detection scenarios, multi-class anomaly detection is not only more prevalent but also more valuable, as it only requires training one model for N classes, whereas single-class methods require training N models for N classes. Thus, this paper focuses on the promising and challenging multi-class anomaly detection.

\begin{figure}[ht]
  \centering
  \includegraphics[width=0.8\textwidth]{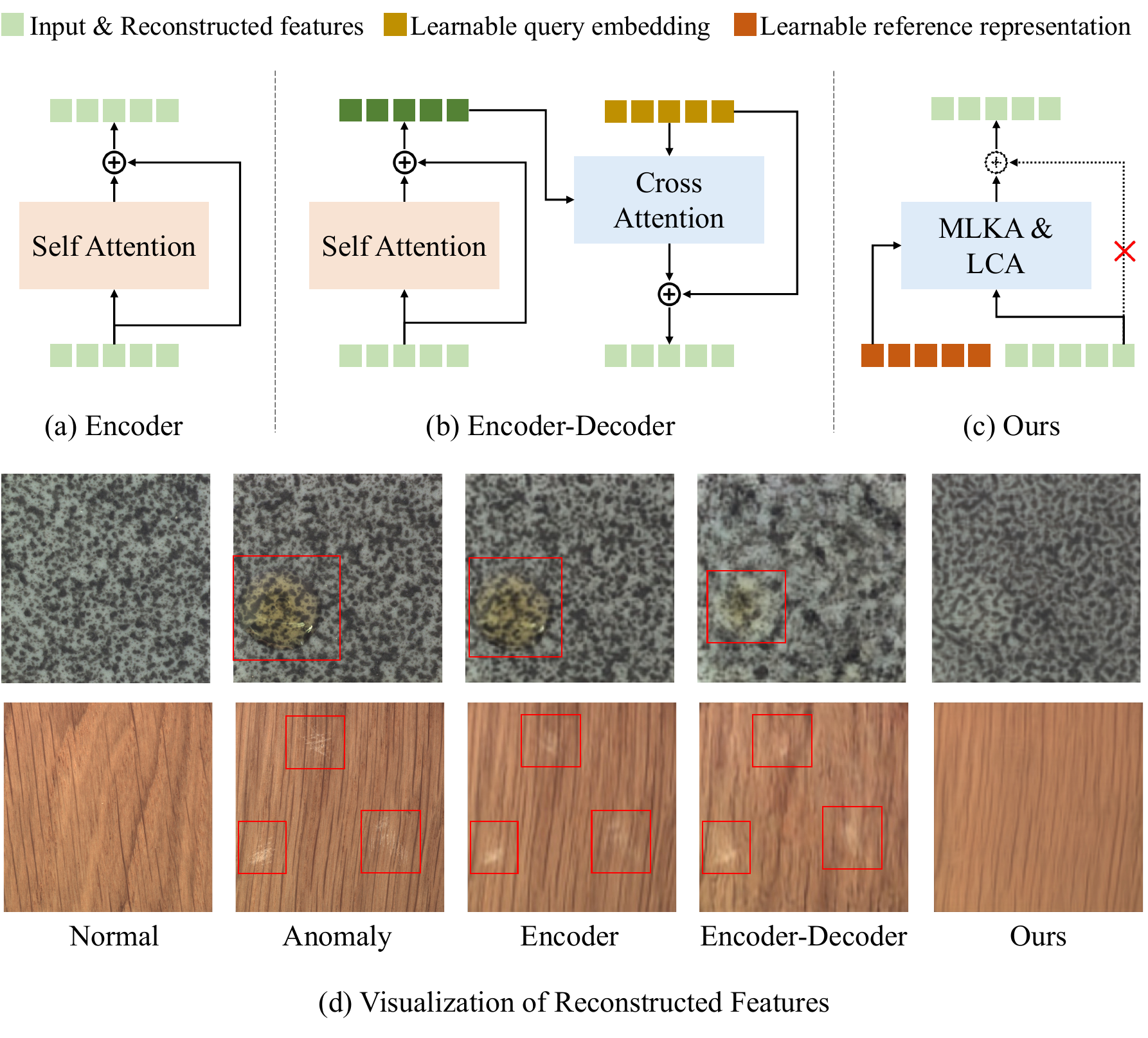}
  \caption{\textbf{Motivation and effectiveness of our method}. Existing frameworks are shown in (a) and (b), fall into learning shortcuts issue. Our framework is depicted in (c), which utilizes learnable reference representation for feature reconstruction to address this issue. (d) shows the visualizations of the reconstructed features, which includes the Normal Sample, the input Anomaly Sample, as well as the features recovered from the Encoder, Encoder-Decoder (specifically UniAD~\cite{you2022unified}) and our proposed methods.}
  \label{fig:intro}
\end{figure}

% 2. limitaion of existing works
Most existing mainstream anomaly detection methods are single-class methods, they can be categorized into embedding-based, knowledge distillation-based, and reconstruction-based approaches. Directly applying single-class method to multi-class anomaly detection leads to significant performance degradation, since the normal patterns for multiple classes is much more complicated than single class. Embedding-based methods~\cite{defard2021padim, rippel2021modeling, roth2022towards, gudovskiy2022cflow, lei2023pyramidflow} employ statistical methods, such as Gaussian distribution~\cite{defard2021padim}, normalizing flow~\cite{gudovskiy2022cflow}, and memory bank~\cite{roth2022towards} to obtain the normal patterns from pre-trained model. However, these methods require high computational or storage resources in multi-class anomaly detection. Knowledge distillation-based methods~\cite{bergmann2020uninformed, salehi2021multiresolution, deng2022anomaly, tien2023revisiting} distill the normal features extracted by a pre-trained teacher model into a student model and assume the student model only learns normal patterns. However, this assumption may fail in multi-class anomaly detection. Reconstruction-based methods~\cite{zavrtanik2021draem, you2022adtr, yan2021learning, yao2023focus, zhang2023diffusionad} train models solely using normal samples, with expectation that the model will learn to recognize normal patterns. Therefore they can reconstruct anomalies as normal instances, enabling the detection of anomalies during inference. However, the fundamental objective of unsupervised reconstruction training is essentially identity mapping or artificial noise eliminating. This carries the risk that the model may not effectively learn normal patterns, but instead encounter the problem of \textit{learning shortcuts} to achieve training objectives, such as identical mapping, resulting in anomalies still being recovered as anomalies.

% 3. limitaion of multi-class AD
Since existing single-class anomaly detection methods cannot be directly applied to multi-class anomaly detection, recent efforts have proposed unified frameworks for multi-class anomaly detection, such as UniAD~\cite{you2022unified} and OmniAL~\cite{zhao2023omnial}. These methods are based on reconstruction, and the tendency of reconstruction models to learn shortcuts during the training phase becomes more pronounced in multi-class anomaly detection. Therefore, they aim to weaken this tendency by increasing the difficulty of learning shortcuts. However, they cannot completely eliminate this possibility because fundamentally, their output is primarily determined by the input, so the model still can learn shortcuts during the training phase to perfectly reconstruct the input.

% 4. motivation
To address this issue, we propose a simple yet effective feature reconstruction method based on learnable reference representation for multi-class anomaly detection. Existing Transformer-based feature reconstruction frameworks, both vanilla Encoder (Figure \ref{fig:intro}(a), like AnoViT~\cite{lee2022anovit}) and Encoder-Decoder (Figure \ref{fig:intro}(b), like UniAD~\cite{you2022unified}), fall into ``learning shortcuts'' since their output is a direct mapping of the input. Therefore they can learn shortcuts such as identity mapping or simple noise elimination, allowing it to perfectly reconstruct features during training without learning the normal patterns. In contrast, our method reconstructs features from learnable reference to ensure our model's output is primarily influenced by the learnable reference. As shown in Figure \ref{fig:intro}(c), the learnable reference representation serve as the Key and Value to recover the features, meanwhile we remove residual connections. Figure \ref{fig:intro}(d) demonstrates the visualization of the reconstructed features, our method presents a higher accuracy in reconstructing anomaly features into normal features.

% 5. Overview method
Specifically, we propose the framework called RLR, which Reconstruct features from Learnable Reference representation. Our RLR is based on Transformer Encoder without residual connections. We utilize two novel components, namely Local Cross Attention (LCA) and Masked Learnable Key Attention (MLKA), to reconstruct features. In the LCA module, we compute the Cross Attention between the input features and the learnable reference. Additionally, considering that the feature maps extracted by the pre-trained CNN model exhibit locality~\cite{roth2022towards}, we introduce a locality constraint for the learnable reference to obtain more effective and accurate reference normal patterns. In the MLKA module, we utilize the learnable reference as the Key to compute neighbor masked Attention without residual connection, allowing us to capture more detailed information to assist in feature reconstruction. This helps prevent the reconstructed features from being overly smoothed when relying solely on LCA.

% 5. experiment
We conducted experiments on two widely used industrial anomaly detection datasets, namely MVTec-AD~\cite{bergmann2019mvtec} and VisA~\cite{zou2022spot}. Our approach outperformed the previous state-of-the-art unified framework and separate anomaly detection models adapted for multi-class task on both datasets. Specifically, we obtain $98.6\%$ Image-AUROC and $98.5\%$ Pixel-AUROC on the MVTec-AD, which is a significant improvement compared to previous methods.

% 6. summary
Overall, our contributions are summarized as follows:
\begin{itemize}
\item We propose a unified anomaly detection framework that reconstructs features from learnable reference representation, thereby forcing the model to learn normal patterns instead of learning shortcuts.
\item We introduce Local Cross Attention to enable the model to learn more accurate and effective reference representation, and Masked Learnable Key Attention to assist the model in reconstructing more detailed features.
\item Our unified anomaly detection framework achieves state-of-the-art performance on popular industrial anomaly detection datasets, MVTec-AD and VisA.
\end{itemize}

\section{Related Work}

\textbf{Unsupervised Anomaly Detection.}~ We divide the mainstream unsupervised anomaly detection approaches into embedding based, knowledge distillation based and reconstruction based methods. The first involves conducting statistical analysis on embedding of normal samples that extracted by pre-trained models. For instance, MDND~\cite{rippel2021modeling} and PaDiM~\cite{defard2021padim} rely on multivariate Gaussian distribution, Patchcore~\cite{roth2022towards} utilizes memory bank, and~\cite{rudolph2021same, gudovskiy2022cflow, rudolph2022fully, lei2023pyramidflow} count on normalizing flow. However, these methods require substantial resources and intricate design tricks. The second distillates the normal features extracted by pre-trained teacher model into student model~\cite{bergmann2020uninformed}. However, despite these approaches require tricks to ensure the student model only learns normal patterns, such as multiresolution~\cite{salehi2021multiresolution} and reverse distillation~\cite{deng2022anomaly, tien2023revisiting}, it still cannot guarantee that. The third applies models like AutoEncoder~\cite{zavrtanik2021draem}, GAN~\cite{goodfellow2020generative, yan2021learning}, Diffusion~\cite{ho2020denoising, zhang2023diffusionad} and Transformer~\cite{you2022adtr} to reconstruct anomaly pixels or features into normal instances. However, existing methods fail to safeguard that the model learns normal patterns instead of shortcuts, while our RLR utilizes learnable reference to guarantee that.

\textbf{Transformer based Anomaly Detection.}~ Recently, Transformer~\cite{vaswani2017attention,zhang2024learning} has gained widespread attention for its ability to model long-range dependencies, leading to its utilization in reconstruction-based anomaly detection methods. InTra~\cite{pirnay2022inpainting} employs a Transformer Encoder to reconstruct masked images. AnoViT~\cite{lee2022anovit} and VT-ADL~\cite{mishra2021vt} use Transformer Encoder to reconstruct features, which are further reconstructed into images using CNNs. ADTR~\cite{you2022adtr} highlights Transformer Encoder-based methods suffer from ``identity mapping'' problem and proposes an Encoder-Decoder architecture for feature reconstruction. However, the outputs of these architectures are still primarily derived from the input mapping, leaving room for potential shortcuts in the learning process. This tendency becomes more pronounced, particularly in the case of multi-class anomaly detection, where the challenge of learning multi-class normal patterns increases, further reinforcing the inclination towards learning shortcuts, while our RLR addresses that.

\textbf{Unified Anomaly Detection.}~ UniAD~\cite{you2022unified} first introduces a unified framework for multi-class anomaly detection and highlights the increased likelihood of encountering the ``learning shortcut'' problem in the unified setting, since learning multi-class normal patterns is more challenging than learning shortcuts. UniAD proposes a solution based on the Transformer Encoder-Decoder structure with Neighbor Mask Attention and Feature Noise to increase the difficulty of the model learning shortcuts. However, it does not fundamentally eliminate the possibility to learn shortcuts as described above. OmniAL~\cite{zhao2023omnial} is an image reconstruction based method. It proposes complex and realistic anomaly synthesis, which successfully work in the multi-class anomaly detection. The primary objective of OmniAL is to enhance the learning of multi-class normal patterns by introducing challenging anomalies that make it difficult for the model to learn shortcuts. However this approach heavily relies on the quality of generated anomalous samples and requires significant prior knowledge and resources. Our RLR reconstructs features from learnable reference, forcing the model learn normal patterns rather than shortcuts.

\section{Method}

\begin{figure}[tb]
  \centering
  \includegraphics[width=0.95\textwidth]{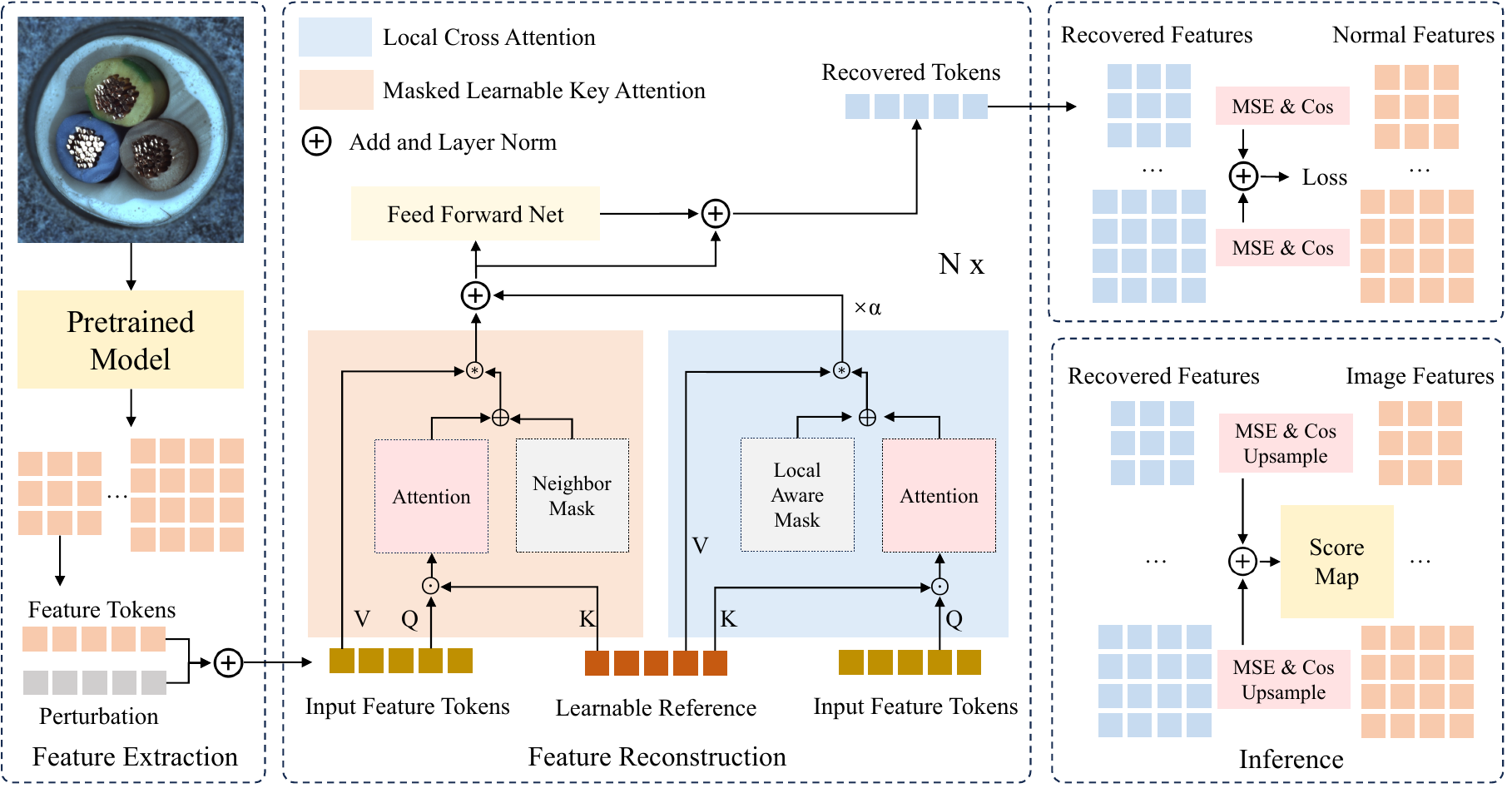}
  \caption{\textbf{Framework of our approach}. RLR consists of Multi-Scale Feature Extraction through pre-trained model, Feature Reconstruction with combination of Masked Learnable Key Attention and Local Cross Attention, Loss and Score Map calculation between recovered features and original features.}
  \label{fig:overview}
\end{figure}

In this paper, we propose a unified anomaly detection framework based on feature reconstruction. The key insight is to reconstruct features from learnable reference representation, thereby enforcing this reference to learn normal patterns. This approach avoids the problem of previous unsupervised reconstruction methods, which employ the self-reconstruction task as the training objective and lead to ``learning shortcuts'' issue. To enhance the model's ability to learn normal reference patterns and reconstruct normal features, we further introduce two simple yet effective attention mechanisms. Figure~\ref{fig:overview} shows the overview of the proposed RLR with four steps: multi-scale feature extraction, feature reconstruction with our parallel Masked Learnable Key Attention and Local Cross Attention, loss calculation between the recovered features and the original features, and score map forecast at the inference phase.

\subsection{Feature Extraction}
\label{sec:extration}

\textbf{Multi-Scale Feature Extraction.}~ Following the previous patch feature based approaches~\cite{defard2021padim, roth2022towards, you2022unified}, we apply the fixed CNNs that pre-trained on ImageNet~\cite{deng2009imagenet} like ResNet~\cite{he2016deep} or EfficientNet~\cite{tan2019efficientnet} to extract the multi-scale feature maps of the input images. The feature maps are defined as $\mathcal{F}_{i,j} \in \mathbb{R}^{C_j \times H_j \times W_j}$, where $i$ is the index of the input image $x_i$, $j$ is the index of the multi-scale feature maps, and $C_j, H_j, W_j$ are the channel dimension, height and width of the $j_{th}$ feature map. Same as~\cite{roth2022towards}, we apply the local neighbourhood aggregation to feature maps to improve robustness. Formally, for a patch at location $(h, w)$, we denote its neighborhood as
\begin{equation}
\begin{split}
  \mathcal{N}^{(h,w)}_p = \left\{(a, b) | a \in [h - \lfloor p/2 \rfloor, ..., h + \lfloor p/2 \rfloor], ~~~~\right. \\
  \left. b \in [w - \lfloor p/2 \rfloor, ..., w + \lfloor p/2 \rfloor] \right\},
  \label{eq:neighbor}
\end{split}
\end{equation}
where $p$ is the neighbor window size. We use Adaptive Average Pooling as the aggregation function $f_{agg}$ to obtain locally aware feature for patch located at $(h, w)$ with its neighborhood $\mathcal{N}^{(h,w)}_p$, which is formulated as
\begin{equation}
  \mathcal{F}_{org,h,w}^{i,j} = f_{agg}\left(\{ \mathcal{F}_{i,j}^{(a, b)} | (a, b) \in \mathcal{N}^{(h,w)}_p \}\right),
  \label{eq:aggregation}
\end{equation}
where $\mathcal{F}_{org}^{i,j} \in \mathbb{R}^{C_j \times H_j \times W_j}$ are the feature maps that are expected to be reconstructed.

\textbf{Feature Perturbation.}~ Using the feature $\mathcal{F}_{org}^{i,j}$ directly as input can lead to the problem of identical mapping due to the presence of residual shortcuts in the vanilla Transformer. Although we have addressed this issue with our shortcut-free attention, feature perturbation is still beneficial for enhancing generalization and robustness. Inspired by~\cite{liu2023simplenet, you2022unified}, we simply sample the noise tokens $\mathcal{P}$ from a Gaussian distribution $\mathcal{N}(0, \sigma^2)$, where $\sigma$ is the adjustable variance. By adding random noise to $\mathcal{F}_{org}^{i,j}$, we define the input feature tokens $\mathcal{F}_{in}^{i,j} \in \mathbb{R}^{H_jW_j \times C_j}$ as 
\begin{equation}
  \mathcal{F}_{in}^{i,j} = Reshape(\mathcal{F}_{org}^{i,j}) + \mathcal{P}.
  \label{eq:fin}
\end{equation}

\subsection{Feature Reconstruction}
\label{sec:recover}

To avoid ``learning shortcut'' issue described above, we replace self-attention in vanilla Transformer with our proposed parallel Masked Learnable Key Attention (MLKA) and Local Cross Attention (LCA). We combine the output of MLKA and LCA, where output of LCA multiplied by a hyperparameter $\alpha$ greater than $1$, to ensure model focuses more on LCA. Note that there is not residual shortcut between the input and the output of the attention module. Hence the new block is stacked by our proposed attention and Feed Forward Network (FFN) with norm \& add similar to vanilla Transformer. For the input feature tokens $\mathcal{F}_{in}^j$ at each feature level $j$, after $K$ consecutive blocks, we obtain the recovered tokens $\mathcal{F}_{out}^j \in \mathbb{R}^{N_j \times C_j}$, where $N_j$ is the number of patches of $j_{th}$ feature map, i.e., $N_j = H_j \times W_j$. We formally define the calculation process of $k_{th}$ block as
\begin{equation}
  \mathcal{Z}_k^j\!=\!LN\left(MLKA(\mathcal{Y}_{k - 1}^j, \mathcal{R}_h^j)\!+\!\alpha LCA(\mathcal{Y}_{k - 1}^j, \mathcal{R}_h^j)\right)
  \label{eq:zk}
\end{equation}
and
\begin{equation}
  \mathcal{Y}_k^j = LN\left(FFN(\mathcal{Z}_k^j) + \mathcal{Z}_k^j\right),
  \label{eq:yk}
\end{equation}
where $\mathcal{Z}_k^j \in \mathbb{R}^{N_j \times C_h}$ is the output of the attention module, $C_h$ means the hidden layer dimension, $LN$ represents LayerNorm, $\mathcal{R}_h^j \in \mathbb{R}^{N_j \times C_h}$ indicates the learnable reference tokens after projected to hidden feature space, and $\mathcal{Y}_k^j \in \mathbb{R}^{N_j \times C_h}$ is the output of the $k_{th}$ block and the input of the $(k + 1)_{th}$ block. We use a single $\mathcal{R}_h^j$ for all $K$ blocks, so it is independent of $k$. Furthermore, we acquire $\mathcal{Y}_0^j$ from $\mathcal{F}_{in}^j$ and $\mathcal{F}_{out}^j$ from $\mathcal{Y}_K^j$, which is defined as
\begin{equation}
  \mathcal{Y}_0^j = FFN(\mathcal{F}_{in}^j), \mathcal{F}_{out}^j = FFN(\mathcal{Y}_K^j).
  \label{eq:inout}
\end{equation}
Then we reshape the $\mathcal{F}_{out}^j$ to the recovered feature maps $\mathcal{F}_{rec}^j \in \mathbb{R}^{C_j \times H_j \times W_j}$ for training and inference.

\textbf{Masked Learnable Key Attention.}~ The residual connection in Transformer allows the output directly contains input, making it easier to learn shortcuts, thus a natural idea to figure it out is to remove this connection. Unfortunately, the skip connection is quite important for training deep self-attention network, as it helps prevent deep Transformer from converging to rank collapse~\cite{he2023deep}. To overcome this challenge, we propose a modified self-attention called Masked Learnable Key Attention (MLKA).

Specifically, we randomly initialize a learnable reference feature representation for each feature level $j$ and define it as $\mathcal{R}^j \in \mathbb{R}^{N_j \times C_j}$, which has the same size as $\mathcal{F}_{in}^j$ because their patch positions need to correspond. Then we use a fully connected layer to project $\mathcal{R}^j$ to the hidden feature space into $\mathcal{R}_h^j \in \mathbb{R}^{N_j \times C_h}$, and a single $\mathcal{R}_h^j$ is used for all K layers. Through training on the normal samples, the learnable reference eventually represents the normal feature pattern. We enhance the vanilla self-attention structure by applying the learnable normal reference feature tokens as the Key vectors. This allows us to eliminate the residual connection and prevent it from falling into rank collapse at the same time. Additionally, inspired by~\cite{you2022unified} we add a neighbor masked attention map to the original attention map between Query and Key vectors in order to make a token invisible to itself and its neighbors when calculating attention. The process of MLKA in $k_{th}$ block is formulated as
\begin{equation}
  \mathcal{Q}_k, \mathcal{K}_k, \mathcal{V}_k = W_k^Q\mathcal{Y}_{k - 1}^j, W_k^K\mathcal{R}_h^j, W_k^V\mathcal{Y}_{k - 1}^j
  \label{eq:mlka1}
\end{equation}
\begin{equation}
  \mathcal{A}_k = Softmax\left(\mathcal{Q}_k\mathcal{K}_k^T/\sqrt{d_k} + \mathcal{M}_{nei}\right)
  \label{eq:mlka2}
\end{equation}
\begin{equation}
  \mathcal{O}_k = \mathcal{A}_k\mathcal{V}_k
  \label{eq:mlka3}
\end{equation}
where parameters $W_k^Q, W_k^K$ and $W_k^V$ are responsible for embedding $\mathcal{Y}_{k - 1}^j$ and $\mathcal{R}_h^j$ into Query, Key and Value vectors, and $\mathcal{M}_{nei} \in \mathbb{R}^{N_j \times N_j}$ is the neighbor mask map that contains zero and negative infinity. In particular, the neighbor patches of the $i_{th}$ patch in $\mathcal{M}_{nei}[i] \in \mathbb{R}^{N_j}$ are marked as negative infinity so that they will be zero after Softmax function, meaning the attention weights of neighbor patches to the $i_{th}$ patch in $\mathcal{A}_k[i] \in \mathbb{R}^{N_j}$ are zero, aiming to ignore patch$_i$'s neighbors. The size of the neighbor window for each feature level $j$ is a hyperparameter, and we provide specific settings in the experimental section. Since MLKA does not have residual connection and with neighbor mask, a token will not directly contain itself after passing through MLKA. Furthermore, due to the learnable key only consists of normal features, the abnormal tokens will receive a low similarity scores when calculating attention. As a result, the output of MLKA will not retain abnormal features. These advantages enable MLKA to recover abnormal tokens to their normal state and prevent the occurrence of ``learning shortcuts''.

\textbf{Local Cross Attention.}~ Although MLKA partially addresses the issue of learning shortcuts, there may still be possibility of residual abnormal features remaining in the output tokens, as the Value vectors correspond to input features, similar to UniAD~\cite{you2022unified}. To overcome this limitation, we propose the Local Cross Attention (LCA) module to reconstruct the feature tokens from learnable reference representation tokens. The basic idea is to treat input feature tokens as Query vectors and learnable reference tokens as Key and Value vectors, and then apply a cross attention between them. However, since the feature maps obtained from CNN-based backbone represent local patch features, incorporating locality constraints can enhance the learning effectiveness of reference. Specifically, the LCA introduces a local aware mask to the attention between input feature and learnable reference, so that a token can only see the reference tokens within a local window to reconstruct its normal representation. The process of LCA can be formulated as
\begin{equation}
  \mathcal{Q}_k, \mathcal{K}_k, \mathcal{V}_k = W_k^Q\mathcal{Y}_{k - 1}^j, W_k^K\mathcal{R}_h^j, W_k^V\mathcal{R}_h^j
  \label{eq:lca1}
\end{equation}
\begin{equation}
  \mathcal{A}_k = Softmax\left(\mathcal{Q}_k\mathcal{K}_k^T/\sqrt{d_k} + \mathcal{M}_{loc}\right)
  \label{eq:lca2}
\end{equation}
\begin{equation}
  \mathcal{O}_k = \mathcal{A}_k\mathcal{V}_k
  \label{eq:lca3}
\end{equation}
where $\mathcal{M}_{loc} \in \mathbb{R}^{N_j \times N_j}$ is the local aware mask similar to $\mathcal{M}_{nei}$ but the neighbor patches are marked as zero while others are negative infinity, meaning the attention weights of non-neighbor patches to the $i_{th}$ patch in $\mathcal{A}_k[i] \in \mathbb{R}^{N_j}$ are zero, thereby focusing attention only on the neighbor patches. Since LCA forces the network to reconstruct the feature tokens from learnable reference representation, the reference eventually contains the normal features, thereby the recovered feature will get rid of abnormal features remarkably.

% \begin{figure}[tb]
%   \centering
%   \includegraphics[width=0.45\textwidth]{img/attention.pdf}
%   \caption{\textbf{Architecture of the Masked Learnable Key Attention and Local Cross Attention}. The Query, Key, Value and Mask tokens are reshaped to 2-dimensional feature map form for intuitive visualization.}
%   \label{fig:attention}
% \end{figure}

\subsection{Training and Loss Function}
\begin{sloppypar}
Following the Feature Reconstruction module, we acquire the recovered multi-scale feature maps $\mathcal{F}_{rec}^j \in \mathbb{R}^{C_j \times H_j \times W_j}$. We utilize Mean Square Error and Cosine Similarity to measure the loss between reconstructed features $\mathcal{F}_{rec}^j$ and the original extracted features $\mathcal{F}_{org}^j$. The loss function is defined as
\end{sloppypar}
\begin{equation}
  \mathcal{L}_{cos}^j = 1 - \frac{1}{H_j W_j} \times \frac{\mathcal{F}_{rec}^j \cdot \mathcal{F}_{org}^j}{\lVert \mathcal{F}_{rec}^j \rVert _2 \lVert \mathcal{F}_{org}^j \rVert _2}
  \label{eq:cos}
\end{equation}
\begin{equation}
  \mathcal{L} = \sum_{j=1}^{L} \left(\frac{\lVert \mathcal{F}_{rec}^j - \mathcal{F}_{org}^j \rVert _2}{H_j W_j} + \mathcal{L}_{cos}^j \right),
  \label{eq:loss}
\end{equation}
where $\mathcal{L}_{cos}$ is the cosine loss and $\mathcal{L}$ is the total loss. After training on the dataset that only contains normal samples, we obtain the network parameters and the reference representation of normal patterns.

\subsection{Inference and Scoring Function}
During inference, we extract the multi-scale feature maps ${F}_{org}^j$ from a test image without introducing any noise. We employ the reconstruction network to recover the corresponding normal feature maps ${F}_{rec}^j$. We also utilize Mean Square Error and Cosine Similarity to achieve the abnormal score map since the recovered feature map ${F}_{rec}^j$ supposed to only contains normal characteristics. The difference between the original and reconstructed features indicates that the original abnormal features have been recovered as normal. The magnitude of this difference serves as indicator of the anomaly score, with a higher difference corresponding to a higher score. We calculate the score map by the following formula
\begin{equation}
  \mathcal{S}_{cos}^j = 1 - \frac{\mathcal{F}_{rec}^j \cdot \mathcal{F}_{org}^j}{\lVert \mathcal{F}_{rec}^j \rVert _2 \lVert \mathcal{F}_{org}^j \rVert _2}
  \label{eq:cosi}
\end{equation}
\begin{equation}
  \mathcal{S} = \frac{\sum_{j=1}^{L} Upsample \left(\lVert \mathcal{F}_{rec}^j - \mathcal{F}_{org}^j \rVert _2 + \mathcal{S}_{cos}^j \right)}{L},
  \label{eq:lossi}
\end{equation}
where $\mathcal{S} \in \mathbb{R}^{H \times W}$ is the final score map. For each feature level $j$, we upsample the $j_{th}$ score map to the original image size, and then calculate the average of all score maps as the final score map.

\section{Experiment}

\subsection{Datasets and Metrics}

To validate the effectiveness of our proposed RLR, we perform experiments on two major industrial anomaly detection datasets MVTec-AD~\cite{bergmann2019mvtec} and VisA~\cite{zou2022spot}. 

\textbf{MVTec-AD}~\cite{bergmann2019mvtec} is a comprehensive industrial anomaly detection dataset that contains 5 texture and 10 object real world industrial products. The training set contains 3,629 high-resolution normal images, while the testing set consists of 467 normal images and 1,258 anomaly images. Each product exhibits multiple types of anomalies with various size, shape, and other morphological attributes.

\textbf{VisA}~\cite{zou2022spot} contains 12 objects and a total of 10,821 images with 9,621 normal samples and 1,200 anomalous samples. It contains 3 subsets consists of complex structure, multiple instance and aligned object. The anomalous images contain various flaws, including surface defects such as scratches, dents, color spots or crack, and structural defects like misplacement or missing parts. We follow the VisA 1-class protocol to do the unsupervised anomaly detection experiments.

\textbf{Metrics.} Follow the previous anomaly detection works~\cite{defard2021padim, roth2022towards, you2022unified, liu2023simplenet}, we use the common metric, Area Under the Receiver Operating Curve (AUROC), to evaluate the performance of the approaches. It contains Image-AUROC for measuring detection performance and Pixel-AUROC for measuring localization performance.

\subsection{Implementation Details}

We train a unified anomaly detection model for all categories. The input images are resized to $256 \times 256$ and features are extracted by EfficientNet-B6~\cite{tan2019efficientnet} that pretrained on ImageNet~\cite{deng2009imagenet}. We reconstruct the feature maps from stage 1 to 3. Respectively the hidden layer dimensions for feature reconstruction are set to 128, 256 and 512, with corresponding neighbor window sizes of 5, 7 and 11. The hyperparameter $\alpha$ is set to 2, and the number of encoder layers in the feature reconstruction module is 4. We train the model using the Adam optimizer for 200 epochs, with an initial learning rate of 1e-4.

\subsection{Main Results}

\begin{table}[t]
\centering
\begin{center}
\caption{\textbf{Quantitative results with SOTA methods on benchmark MVTec-AD}. Anomaly detection and localization results are displayed as Image-AUROC\% / Pixel-AUROC\%. The best results are highlighted in \textbf{bold}.}
\resizebox{\textwidth}{!}{
  \begin{tabular}{c|c|c c c c c c c|c}
  \toprule
  \multicolumn{2}{c|}{Category} & PaDiM~\cite{defard2021padim} & MKD~\cite{salehi2021multiresolution} & DRAEM~\cite{zavrtanik2021draem} & Patchcore~\cite{roth2022towards} & SimpleNet~\cite{liu2023simplenet} & UniAD~\cite{you2022unified} & OmniAL~\cite{zhao2023omnial} & Ours \\
  \midrule
  \multirow{10}*{\rotatebox{-90}{Object~~}}
    & Bottle     & 97.9 / 96.1 & 98.7 / 91.8 & 97.5 / 87.6 & \textbf{100} / 98.4 & 86.5 / 88.1 & 99.7 / 98.1 & \textbf{100} / \textbf{99.2}  & \textbf{100} / 99.0  \\
    & Cable      & 70.9 / 81.0 & 78.2 / 89.3 & 57.8 / 71.3 & 99.2 / 97.3 & 71.5 / 79.3 & 95.2 / 97.3 & 98.2 / 97.3 & \textbf{99.3} / \textbf{99.0} \\
    & Capsule    & 73.4 / 96.9 & 68.3 / 88.3 & 65.3 / 50.5 & 85.6 / 95.2 & 77.8 / 89.4 & 86.9 / 98.5 & 95.2 / 96.9 & \textbf{97.0} / \textbf{99.2} \\
    & Hazelnut   & 85.5 / 96.3 & 97.1 / 91.2 & 93.7 / 96.9 & \textbf{100} / 98.9 & 94.3 / 95.9 & 99.8 / 98.1 & 95.6 / 98.4 & \textbf{100} / \textbf{99.0}  \\
    & Metal Nut  & 88.0 / 84.8 & 64.9 / 64.2 & 72.8 / 62.2 & \textbf{99.9} / 98.4 & 87.8 / 87.0 & 99.2 / 94.8 & 99.2 / \textbf{99.1} & 99.7 / 98.3 \\
    & Pill       & 68.8 / 87.7 & 79.7 / 69.7 & 82.2 / 94.4 & 93.3 / 95.7 & 80.2 / 90.7 & 93.7 / 95.0 & 97.2 / \textbf{98.9} & \textbf{98.9} / 98.3 \\
    & Screw      & 56.9 / 94.1 & 75.6 / 92.1 & 92.0 / 95.5 & 82.9 / 95.9 & 72.8 / 85.7 & 87.5 / 98.3 & 88.0 / 98.0 & \textbf{94.8} / \textbf{99.5} \\
    & Toothbrush & 95.3 / 95.6 & 75.3 / 88.9 & 90.6 / 97.7 & 88.9 / 98.2 & 87.8 / 96.4 & 94.2 / 98.4 & \textbf{100} / \textbf{99.4}  & 93.1 / 98.9 \\
    & Transistor & 86.6 / 92.3 & 73.4 / 71.7 & 74.8 / 64.5 & 96.7 / 89.3 & 79.7 / 83.3 & \textbf{99.8} / 97.9 & 93.8 / 93.3 & 99.7 / \textbf{98.6} \\
    & Zipper     & 79.7 / 94.8 & 87.4 / 86.1 & 98.8 / 98.3 & 91.9 / 95.5 & 88.5 / 84.3 & 95.8 / 96.8 & \textbf{100} / \textbf{99.5}  & 98.5 / 98.2 \\
  \midrule
  \multirow{5}*{\rotatebox{-90}{Texture~~~}}
    & Carpet     & 93.8 / 97.6 & 69.8 / 95.5 & 98.0 / 98.6 & 96.1 / 98.7 & 87.6 / 89.5 & \textbf{99.8} / 98.5 & 98.7 / \textbf{99.4} & 99.7 / 99.0 \\
    & Grid       & 73.9 / 71.0 & 83.8 / 82.3 & 99.3 / 98.7 & 97.1 / 96.6 & 79.1 / 69.9 & 98.2 / 96.5 & \textbf{99.9} / \textbf{99.4} & 99.8 / 98.7 \\
    & Leather    & 99.9 / 84.8 & 93.6 / 96.7 & 98.7 / 97.3 & \textbf{100} / \textbf{99.4} & 95.2 / 96.6 & \textbf{100} / 98.8  & 99.0 / 99.3 & \textbf{100} / \textbf{99.4}  \\
    & Tile       & 93.3 / 80.5 & 89.5 / 85.3 & 99.8 / 98.0 & 99.9 / 95.7 & 97.9 / 91.6 & 99.3 / 91.8 & 99.6 / \textbf{99.0} & \textbf{100} / 96.7  \\
    & Wood       & 98.4 / 89.1 & 93.4 / 80.5 & 99.8 / 96.0 & 98.4 / 93.5 & 97.5 / 87.0 & 98.6 / 93.2 & 93.2 / \textbf{97.4} & \textbf{98.9} / 95.5 \\
  \midrule
  \multicolumn{2}{c|}{Mean} & 84.2 / 89.5 & 81.9 / 84.9 & 88.1 / 87.2 & 95.3 / 96.4 & 85.6 / 87.6 & 96.5 / 96.8 & 97.2 / 98.3 & \textbf{98.6} / \textbf{98.5} \\
  \bottomrule
  \end{tabular}
}
\label{table:mvtec}
\end{center}
\end{table}

\begin{table}[!h]
\begin{center}
\caption{\textbf{Quantitative results with SOTA methods on benchmark VisA}. Anomaly detection and localization results are displayed as Image-AUROC\%/Pixel-AUROC\%. The best results are highlighted in \textbf{bold}.}
\resizebox{\textwidth}{!}{
  \begin{tabular}{c|c|c c c c|c}
  \toprule
  \multicolumn{2}{c|}{Category} & DRAEM~\cite{zavrtanik2021draem} & JNLD~\cite{zhao2022just} & OmniAL~\cite{zhao2023omnial} & UniAD~\cite{you2022unified} & Ours \\
  \midrule
  \multirow{4}{*}{\makecell[c]{Complex \\ structure}}
    & PCB1        & 83.9 / 94.0 & 82.9 / 98.0 & 77.7 / 97.6 & 94.2 / 99.4 & \textbf{97.0} / \textbf{99.7} \\
    & PCB2        & 81.7 / 94.1 & 79.1 / 95.0 & 81.0 / 93.9 & 93.3 / 97.8 & \textbf{97.4} / \textbf{99.1} \\
    & PCB3        & 87.7 / 94.1 & 90.1 / 98.5 & 88.1 / 94.7 & 87.2 / 98.2 & \textbf{96.4} / \textbf{99.1} \\
    & PCB4        & 87.1 / 72.3 & 96.2 / 97.5 & 95.3 / 97.1 & 99.2 / 97.8 & \textbf{99.7} / \textbf{98.4} \\
  \midrule
  \multirow{4}{*}{\makecell[c]{Multiple \\ instance}}
    & Macaroni1   & 68.6 / 89.8 & 90.5 / 93.3 & 92.6 / 98.6 & 91.6 / 99.2 & \textbf{97.7} / \textbf{99.8} \\
    & Macaroni2   & 60.3 / 83.2 & 71.3 / 92.1 & 75.2 / 97.9 & 83.9 / 97.9 & \textbf{86.3} / \textbf{99.3} \\
    & Capsules    & 89.6 / 96.6 & 91.4 / \textbf{99.6} & \textbf{90.6} / 99.4 & 73.1 / 98.1 & 85.9 / 99.2 \\
    & Candles     & 70.2 / 82.6 & 85.4 / 94.5 & 86.8 / 95.8 & 96.9 / 99.1 & \textbf{98.2} / \textbf{99.5} \\
  \midrule
  \multirow{4}{*}{\makecell[c]{Aligned \\ object}}
    & Cashew      & 67.3 / 68.5 & 82.5 / 94.1 & 88.6 / 95.0 & 93.2 / 98.6 & \textbf{96.1} / \textbf{99.4} \\
    & Chewing gum & 90.0 / 92.7 & 96.0 / 98.9 & 96.4 / 99.0 & 99.0 / 99.1 & \textbf{99.8} / \textbf{99.2} \\
    & Fryum       & 86.2 / 83.2 & 91.9 / 90.0 & 94.6 / 92.1 & 89.3 / \textbf{97.5} & \textbf{96.5} / 97.4 \\
    & Pipe fryum  & 87.1 / 72.3 & 87.5 / 92.5 & 86.1 / 98.2 & 97.3 / 99.1 & \textbf{99.4} / \textbf{99.5} \\
  \midrule
  \multicolumn{2}{c|}{Mean} & 80.5 / 87.0 & 87.1 / 95.2 & 87.8 / 96.6 & 91.5 / 98.5 & \textbf{95.9} / \textbf{99.2} \\
  \bottomrule
  \end{tabular}
}
\label{table:visa}
\end{center}
\end{table}

We compared our proposed method RLR with several previous state-of-the-art (SOTA) anomaly detection approaches, including unified frameworks such as UniAD~\cite{you2022unified} and OmniAL~\cite{zhao2023omnial}, as well as the SOTA separate methods under multi-class anomaly detection setting, such as MDK~\cite{salehi2021multiresolution}, DRAEM~\cite{zavrtanik2021draem}, PatchCore~\cite{yi2020patch}, SimpleNet~\cite{liu2023simplenet} and so on. We conducted these comparisons on the MVTec-AD and VisA datasets, and our method outperforms these methods in both anomaly detection and localization metrics, achieving new SOTA results. The specific results of MVTec-AD are presented in Table~\ref{table:mvtec} and VisA in Table~\ref{table:visa}.

\textbf{Anomaly Detection.}~ Our method has achieved SOTA results on both the MVTec-AD and VisA datasets in terms of the anomaly detection metric Image-AUROC (I-A). Furthermore, our approach has shown significant improvements compared to the previous unified framework. In contrast to the feature reconstruction-based method UniAD~\cite{you2022unified}, our method has increased the I-A score from $96.5\%$ to $98.6\%$ on MVTec-AD and $91.5\%$ to $95.9\%$ on VisA. On the other hand, separate models such as PaDiM~\cite{defard2021padim} and DRAEM~\cite{zavrtanik2021draem} have exhibited a notable decrease in performance, indicating that these models struggle to effectively learn normal patterns across multiple classes.

\begin{figure}[t]
  \centering
  \includegraphics[width=0.95\textwidth]{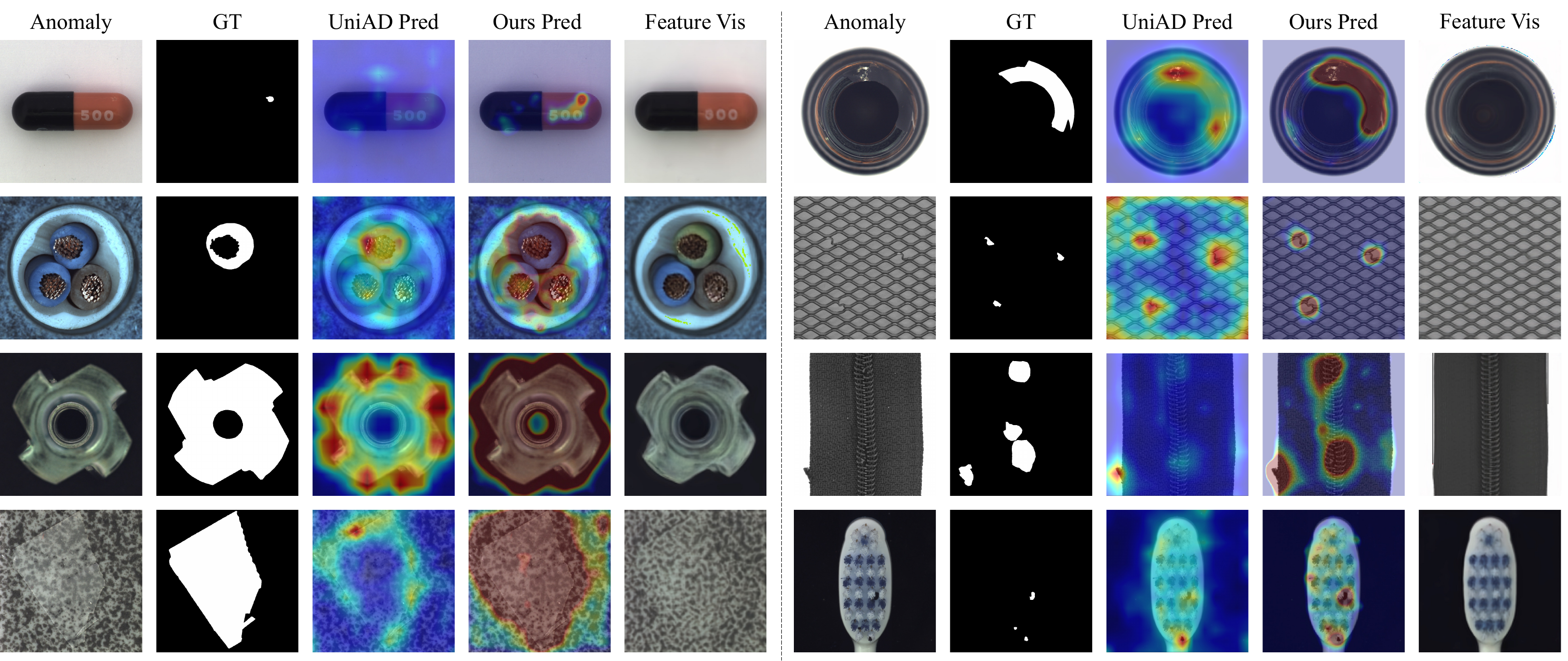}
  \caption{\textbf{Qualitative results on MVTec-AD}. We visualize several anomalies (Anomaly) along with their corresponding Ground Truth (GT), the detection results of UniAD (UniAD Pred), the detection results of our method (Ours Pred), and the visualization of ours reconstructed features (Feature Vis).}
  \label{fig:res_vis}
\end{figure}

\textbf{Anomaly Localization.}~ We have also achieved SOTA results on the pixel-level localization metric Pixel-AUROC (P-A), on both datasets. Since OmniAL~\cite{zhao2023omnial} is a pixel-level reconstruction and detection model, it has an advantage in anomaly localization. However, our method, with its more accurate detection performance, has ultimately improved the P-A score of OmniAL on the MVTec-AD from $98.3\%$ to $98.5\%$, and on the VisA from $96.6\%$ to $99.2\%$. Moreover, compared to the feature reconstruction-based UniAD, our improvements are even more significant on MVTec-AD.

\textbf{Qualitative Results on MVTec-AD.}~ To further demonstrate the superiority of our method, we visualized the reconstruction and detection results of the MVTec-AD~\cite{bergmann2019mvtec} dataset, as shown in Figure~\ref{fig:res_vis}. It can be observed that UniAD fails to detect some anomalies as these anomaly features are not recovered to normal features. In contrast, our method successfully reconstructs normal features. This indicates that UniAD has not fully learned the normal patterns for these classes but instead learned shortcuts that prevent the reconstruction of anomalies as normal features. In contrast, our method recovers these samples to normal instances, demonstrates the effectiveness of our RLR in successfully addressing the issue of models learning shortcuts. We also conducted qualitative analysis on the VisA~\cite{zou2022spot} dataset. Due to space limitations, please refer to the supplementary materials for details.

% \begin{figure}[t]
%   \centering
%   \includegraphics[width=0.95\textwidth]{img/visa.pdf}
%   \caption{\textbf{Qualitative results on VisA}. We visualize several anomalies (Anomaly) along with their corresponding Ground Truth (GT), the detection results of UniAD (UniAD Pred), and the detection results of our method (Ours Pred).}
%   \label{fig:visa}
% \end{figure}

% \textbf{Qualitative Results on VisA.}~We also conducted a qualitative analysis on the VisA~\cite{zou2022spot} dataset. Figure~\ref{fig:visa} presents the visual detection results of UniAD~\cite{you2022unified} and our RLR on the VisA dataset. It is important to note that we were unable to reproduce the results of OmniAL~\cite{zhao2023omnial} due to the unavailability of its source code, hence its results are not shown. Similar to the qualitative analysis conducted on the MVTec-AD~\cite{bergmann2019mvtec} dataset, our method successfully detects samples that UniAD fails to detect, thus demonstrating the effectiveness of our approach.

\subsection{Ablation Study}

\textbf{Ablations of Each Component.}~ We conducted further ablation experiments in the unified setting on the MVTec-AD dataset to validate the effectiveness of each proposed module. The results are shown in Table~\ref{table:ablation}, where each module only operates on the attention mechanism. ``Residual'' indicates the presence of residual connections, ``Self Att.'' computes the self-attention of input features, while in ``Cross Att.'', the Query vector is the input feature, and the Key and Value vectors are learnable reference features.

Firstly, residual connections lead Transformer Encoder-based reconstruction models into the trap of learning shortcuts, whether utilizing Self Attention or Cross Attention. Secondly, the Cross Attention without residual connections in Table~\ref{table:ablation} can represent the foundational idea (baseline) of the our proposed method, which forces the model to reconstruct from learnable reference features explicitly, and we can see that it achieved decent detection performance. Compared with Self Attention with residual connections, its I-A and P-A metrics improved by 12.2\% and 11.8\%, demonstrating the effectiveness of our framework. Additionally, our proposed MLKA and LCA have further improved the metrics I-A and P-A by 2.9\% and 1.4\% compared with Cross Attention, which also proves the effectiveness of the two modules we proposed. Finally, incorporating local constraints in LCA yields better results compared to vanilla Cross Attention with or without MLKA. This indicates that adding local constraints enables the model to learn more effective references.

\begin{table}[t]
\begin{center}
\caption{\textbf{Ablation study of each component on MVTec-AD}. Self Att. means vanilla Self Attention and Cross Att. means Cross Attention with learnable reference representation.}
  \begin{tabular}{c c c c c c c}
  \toprule
  \multicolumn{5}{c}{Modules} & \multicolumn{2}{c}{Metrics} \\
  \cmidrule(lr){1-5} \cmidrule(lr){6-7}
  Residual & Self Att. & Cross Att. & MLKA & LCA & I-A & P-A \\
  \midrule
  \checkmark & \checkmark & ~ & ~ & ~ & 83.5 & 85.3 \\
  \checkmark & ~ & \checkmark & ~ & ~ & 85.6 & 86.9 \\
  ~ & ~ & \checkmark & ~ & ~ & 95.7 & 97.1 \\
  ~ & ~ & ~ & \checkmark & ~ & 96.4 & 97.5 \\
  ~ & ~ & ~ & ~ & \checkmark & 97.9 & 98.0 \\
  ~ & ~ & \checkmark & \checkmark & ~ & 96.8 & 97.9 \\
  ~ & ~ & ~ & \checkmark & \checkmark & \textbf{98.6} & \textbf{98.5} \\
  \bottomrule
  \end{tabular}
\label{table:ablation}
\end{center}
\end{table}

\textbf{Analysis of Multi-Scale Feature Extraction.}~ We investigate the results of combining feature maps at different scales, as shown in Table ~\ref{table:scale}. Since the feature map of stage 4 is too small and no longer provides meaningful reconstruction information, we combine the feature maps from stages 1 to 3. From the results, it can be observed that the feature maps from stages 2 and 3 exhibit better detection performance (I-A). The feature map from stage 1 enhances the ability for anomaly localization (P-A) but may have a slight negative impact on detection (I-A).

\begin{table}[t]
\begin{center}
\caption{\textbf{Different combination choices of multi-scale feature maps on MVTec-AD}. Metrics are presented in the form of Image-AUROC\% / Pixel-AUROC\%.}
  \begin{tabular}{c|c c c}
  \toprule
  Combination & stage 1, 2 & stage 2, 3 & stage 1, 2, 3 \\
  \midrule
  Metrics & 97.6 / 97.9 & \textbf{98.9} / 98.1 & 98.6 / \textbf{98.5} \\
  \bottomrule
  \end{tabular}
\label{table:scale}
\end{center}
\end{table}

\textbf{Analysis of Hyperparameter $\boldsymbol{\alpha}$.}~ We conducted experiments with different values of $\alpha$, and the results are shown in Table~\ref{table:alpha}. It can be observed that alpha values greater than 1 outperform $\alpha$ equals 1. This indicates that MLKA still has the potential for learning shortcuts, therefore it is necessary to pay more attention on LCA. However, excessively high $\alpha$ value can render MLKA ineffective, resulting in a decline in metrics. Additionally, all $\alpha$ values exhibit high metrics, indicating the robustness of our approach.

\begin{table}[t]
\begin{center}
\caption{\textbf{Different parameter $\alpha$ on MVTec-AD}. Metrics are presented in the form of Image-AUROC\% / Pixel-AUROC\%.}
  \begin{tabular}{c|c c c}
  \toprule
  $\alpha = $ & 1 & 2 & 3 \\
  \midrule
  Metrics & 98.3 / 98.4 & \textbf{98.6} / \textbf{98.5} & 98.5 / 98.4 \\
  \bottomrule
  \end{tabular}
\label{table:alpha}
\end{center}
\end{table}

\textbf{Analysis of Feature Perturbation.}~ Since our method has already eliminated the possibility of the model relying on shortcuts, adding noise to the input is not necessary. However, perturbations can still enhance the model's reconstruction ability to some extent. After removing the perturbations, our method achieved a I-A / P-A result of $98.3\% / 98.2\%$ on the MVTec-AD, which is only $0.3\%$ lower than when noise was added in both I-A and P-A metrics. In contrast, in the context of UniAD, perturbations are used to increase the difficulty of the model learning shortcuts, making them more important in the UniAD framework.

\section{Conclusion}
In this paper, we propose the RLR framework to address the issue of learning shortcuts in feature-based reconstruction for multi-class anomaly detection methods. The RLR framework Reconstructs features from Learnable Reference representaion. To further improve the accuracy and effectiveness of the references, as well as the precision and detail of the reconstructed features, we design the Local Cross Attention module and the Mask Learnable Key Attention module. Our experimental results demonstrate the state-of-the-art (SOTA) performance of our method on various datasets. Both qualitative and quantitative analysis validate the effectiveness of our approach.

\section*{Acknowledgements}
This work was supported by National Natural Science Foundation of China, No.62171139.

\bibliographystyle{splncs04}
\bibliography{main}

\newpage

\setcounter{page}{1}
\setcounter{section}{0}
\setcounter{figure}{0}
\setcounter{table}{0}

\onecolumn{
\centering
\Large
\textbf{Learning Unified Reference Representation for Unsupervised Multi-class Anomaly Detection}\\
Supplementary Material \\
}

\section{Overview}

This supplementary material consists of:
\begin{enumerate}
  \item The ablation study of fusion approaches of MLKA and LCA. (Sec.~\ref{sec:fusion})
  \item The qualitative results on VisA dataset. (Sec.~\ref{sec:visa})
  \item The ablation study of the quantity of learnable references. (Sec.~\ref{sec:references})
  \item The failure analysis. (Sec.~\ref{sec:failure})
\end{enumerate}

\section{Fusion of MLKA and LCA}
\label{sec:fusion}

In order to integrate the outputs of MLKA and LCA modules, we explored alternative fusion methods. In addition to employing the hyperparameter $\alpha$ to perform a weighted summation of their outputs, we conducted experiments by introducing learnable parameters to facilitate the weighted summation of the two modules. This modification involved transforming Equation (4) in the main text into
\begin{equation}
\begin{split}
  \mathcal{Z}_k = LN\left(\theta_1 ~MLKA(\mathcal{Y}_{k - 1}, \mathcal{R}_h) \right. \\
  \left. + ~\theta_2 ~LCA(\mathcal{Y}_{k - 1}, \mathcal{R}_h)\right),
  \label{eq:zk2}
\end{split}
\end{equation}
where $\theta_1$ and $\theta_2$ are the parameters learned by the model. The experimental results are presented in Table~\ref{table:at}, which demonstrate that the use of learnable weights increases the likelihood of the model learning shortcuts, potentially focusing more on MLKA rather than LCA. This suggests that by using the hyperparameter $\alpha$, we can encourage the model to pay more attention to LCA, thereby compelling it to learn normal patterns and achieve better detection performance.

\begin{table}[ht]
\begin{center}
\caption{\textbf{Different fusion methods of MLKA and LCA on MVTec-AD}. Metrics are presented in the form of Image-AUROC\% / Pixel-AUROC\%.}
  \begin{tabular}{c|c c}
  \toprule
  Fusion & w/ $\theta$ & w/ $\alpha$ \\
  \midrule
  Metrics & 98.1 / 98.2 & \textbf{98.6} / \textbf{98.5} \\
  \bottomrule
  \end{tabular}
\label{table:at}
\end{center}
\end{table}

\section{Qualitative Results on VisA.}
\label{sec:visa}

\begin{figure}[ht]
  \centering
  \includegraphics[width=0.95\textwidth]{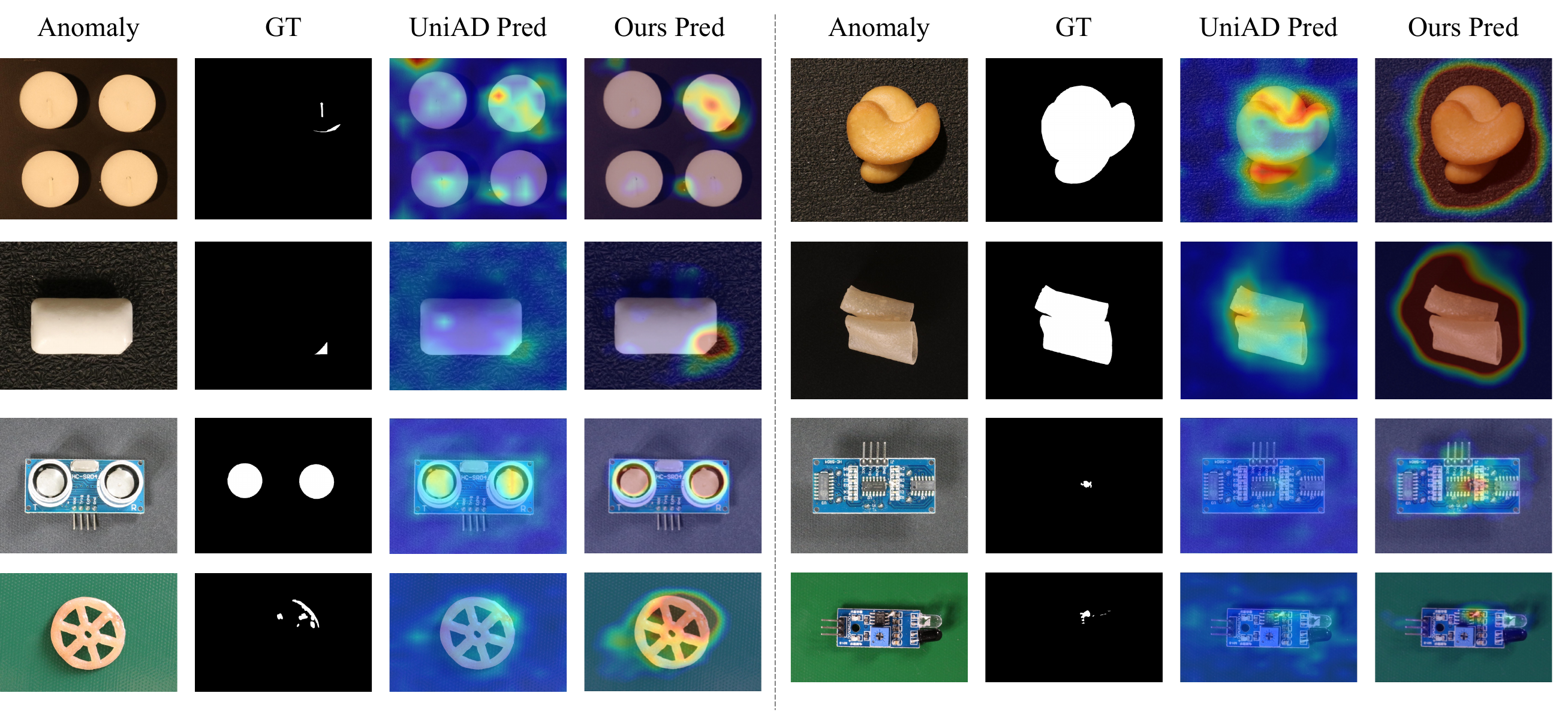}
  \caption{\textbf{Qualitative results on VisA}. We visualize several anomalies (Anomaly) along with their corresponding Ground Truth (GT), the detection results of UniAD (UniAD Pred), and the detection results of our method (Ours Pred).}
  \label{fig:visa}
\end{figure}

To further validate the superiority of our approach, we also conducted a qualitative analysis on the VisA~\cite{zou2022spot} dataset. Figure~\ref{fig:visa} presents the visual detection results of UniAD~\cite{you2022unified} and our RLR on the VisA dataset. It is important to note that we were unable to reproduce the results of OmniAL~\cite{zhao2023omnial} due to the unavailability of its source code, hence its results are not shown. Similar to the qualitative analysis conducted on the MVTec-AD~\cite{bergmann2019mvtec} dataset, our method successfully detects samples that UniAD fails to detect, thus demonstrating the effectiveness of our approach.

\section{Multiple Learnable References}
\label{sec:references}

We conducted experimental analysis on the quantity of learnable references. Specifically, we performed calculations of attention with multiple references in the LCA and summed the outputs, which transformed $LCA(\mathcal{Y}_{k - 1}, \mathcal{R}_h)$ in Equation (4) in the main text into $\sum_{i=1}^{K}LCA(\mathcal{Y}_{k - 1}, \mathcal{R}_h^{i})$, where $K$ is the number of learnable references. Additionally, we imposed constraints on the diversity of references in the loss function. The experimental results are presented in Table~\ref{table:mlr}, which indicates that this approach does not lead to performance improvement; instead, it incurs higher computational and storage resource consumption. Therefore, this paper only utilizes a single learnable reference representation ($K = 1$).

\begin{table}[ht]
\begin{center}
\caption{\textbf{The ablation study results of different $K$ on MVTec-AD}. Metrics are presented in the form of Image-AUROC\% / Pixel-AUROC\%.}
  \begin{tabular}{c|c c c}
  \toprule
  $K = $ & 1 & 2 & 3 \\
  \midrule
  Metrics & \textbf{98.6} / \textbf{98.5} & \textbf{98.6} / \textbf{98.5} & \textbf{98.6} / 98.4 \\
  \bottomrule
  \end{tabular}
\label{table:mlr}
\end{center}
\end{table}

\section{The Failure Analysis}
\label{sec:failure}

We conducted a failure analysis on classes with low performance metrics on the MVTec-AD and VisA datasets. Specifically, in MVTec-AD, the Toothbrush class exhibits lower metrics due to the recognition of noise in the background as anomalies, resulting in some normal samples being misclassified as abnormal samples. This phenomenon is more likely to occur in feature reconstruction methods, as evidenced by similarly low metrics for this class in UniAD. Additionally, in the VisA dataset, the Capsules class exhibits lower metrics owing to the presence of random discolorations. Reconstruction models tend to confuse these discolorations with subtle anomalies, leading to poor performance of RLR and UniAD on this class.

\end{document}